\documentclass[journal]{IEEEtran}

\ifCLASSINFOpdf
\usepackage{graphicx}
\else
\usepackage[dvips]{graphicx}
\fi
\usepackage{amsmath,amssymb}
\usepackage{url}
\usepackage{cite}
\usepackage{tikz}
\usetikzlibrary{shapes.geometric, arrows.meta, positioning, calc}

\usepackage[hidelinks]{hyperref}
\hypersetup{
    colorlinks=false,
    pdfborder={0 0 0},
    breaklinks=true
}

\begin{document}

\title{Hazedefy: A Lightweight Real-Time Image and Video Dehazing Pipeline for Practical Deployment}

\author{
\IEEEauthorblockN{Ayush Bhavsar \\}
\IEEEauthorblockA{
Information Technology\\
NIT Raipur\\
Raipur, India\\
ayush21bhavsar@gmail.com
}

\thanks{Department of Information Technology, National Institute of Technology, Raipur, India.}}
\markboth{}{}
\maketitle

\begin{abstract}
This paper introduces Hazedefy, a lightweight, application-focused dehazing pipeline intended for real-time video and live camera feed enhancement.  Hazedefy prioritizes computational simplicity and practical deployability on consumer-grade hardware, building upon the Dark Channel Prior (DCP) concept and the atmospheric scattering model.  Important elements include gamma-adaptive reconstruction, a fast transmission approximation with lower bounds for numerical stability, a stabilized atmospheric-light estimator based on fractional top-pixel averaging, and an optional color-balance stage.  Reproducible assets and the entire open source implementation are publicly available on Zenodo (DOI: 10.5281/zenodo.17915355)\cite{ayushbhavsar}. An interactive live demonstration of the proposed pipeline is available online.\footnote{https://hazedefy1.netlify.app/}\label{fn1} The pipeline is appropriate for mobile and embedded applications because experimental demonstrations on real-world videos demonstrate perceptually enhanced visibility and contrast without requiring GPU acceleration. 
\end{abstract}


\section{Introduction}
\IEEEPARstart{A}{tmospheric} haze, fog, and pollution scatter light and reduce contrast in outdoor images, decreasing visibility for both humans and cameras. Traditional single image dehazing methods that rely on physical scattering models are easy to understand, but they often run slowly or lose accuracy because of their heavy computations.

The Dark Channel Prior (DCP) \cite{he2011single} is still a common baseline for haze removal because it follows clear physical principles. Latest learning based methods can produce very good looking results, but they usually need GPUs (Graphics Processing Units) and a lot of training data. Due to these requirements, they are not suitable for use on small, embedded, or lightweight devices.

Hazedefy is an application focused dehazing pipeline designed to meet these deployment requirements. It supports two modes: (i) a real-time video mode that runs efficiently on a CPU, and (ii) an offline image mode that uses guided-filter refinement for higher-quality results. The complete codebase is publicly available on Zenodo (DOI: 10.5281/zenodo.17915355)\cite{ayushbhavsar}, and an interactive live demo is available online (https://hazedefy1.netlify.app/).\hyperref[fn1]{\footnotemark[\value{footnote}]}

\section{Related Work}
Research on image and video dehazing includes methods based on physical models, fast filtering techniques, and more recent learning based approaches. This section briefly reviews key works from these areas that are most relevant to Hazedefy..

\subsection{Physics Based Model}
Physics based dehazing methods are based on the atmospheric scattering model, which describes visibility loss as the combined effect of scene radiance weakening and added airlight \cite{narasimhan2002vision}. Among these methods, the Dark Channel Prior (DCP) proposed by He \emph{et al.} \cite{he2011single} is one of the most widely used approaches because it is physically intuitive and works well for single images.

However, traditional DCP based techniques often depend on computationally heavy steps such as soft matting or large kernel filtering. This makes them difficult to use in real time or on devices with limited resources. Other physics based ideas, such as color line constraints \cite{fattal2008single} and contrast based priors \cite{tan2008visibility}, try to reduce the strict assumptions of DCP, but they still suffer from issues like sensitivity to parameters, reduced robustness, or high runtime cost.

These challenges have led to the development of simpler and more stable physicsbased pipelines that keep the benefits of physical understandability while running faster. Hazedefy follows this approach by preserving the core scattering model and the intuition behind DCP, while using lightweight approximations and stability focused design choices that make real-time deployment practical.

\subsection{Fast and Filter Based Methods}
To reduce the computational cost of classical dehazing pipelines, various works have tried fast filtering and refinement strategies. Edge preserving filters, such as guided filtering \cite{he2013guided}, are often used to refine coarse transmission maps while avoiding halo artifacts. Compared to soft matting, these methods are much faster, but they still add extra processing per frame, which can limit real-time performance on CPU based systems.

Other methods focus on simplifying the haze formation assumptions to enable faster inference. Tarel and Hauti\`ere proposed a highly efficient visibility restoration technique suitable for practical systems \cite{tarel2009fast}, while Zhu \emph{et al.} introduced the color attenuation prior (CAP) to estimate transmission using a lightweight depth model \cite{zhu2015learning}. Although these techniques improve runtime efficiency, they often rely on scene-dependent assumptions or require careful parameter tuning.

Hazedefy adopts a similar idea of reducing computational complexity but emphasizes frame-to-frame stability and predictable runtime behavior, making it suitable for live video and embedded deployments.

\subsection{Learning Based Approaches}
Recent learning based dehazing methods use Convolutional Neural Networks (CNN) and end-to-end models to learn complex mappings for haze removal. Approaches such as DehazeNet \cite{cai2016dehazenet}, along with later fusion based and attention driven networks \cite{ren2016gated}, achieve strong visual quality and show good robustness across a wide range of scenes.

Despite their effectiveness, these methods usually depend on GPU acceleration, large labeled datasets, and extensive training. As a result, deploying them on resource limited or real-time systems is often difficult due to large model sizes, slow inference, and hardware requirements.

In contrast, Hazedefy does not rely on data driven training. Instead, it combines interpretable physical priors with lightweight approximations, allowing it to run efficiently on CPU only platforms while maintaining real-time performance.

\subsection{Placement of Hazedefy}
Hazedefy sits at the intersection of physically motivated priors and engineering optimizations for real-time deployment. It adopts DCP style estimation refined for stability and speed (fractional top pixel averaging, clamped transmission, optional guided filtering) and exposes configurations suitable for CPU only, embedded, or mobile usage. This positions Hazedefy as a practical, reproducible baseline for application driven dehazing.

\section{Atmospheric Scattering Model and Notation}
We adopt the common atmospheric scattering model:
\begin{equation}
I(x) = J(x)\,t(x) + A\big(1 - t(x)\big),
\label{eq:scattering}
\end{equation}
where $I(x)$ is the observed hazy image, $J(x)$ is the scene radiance to be recovered, $A$ is the global atmospheric light vector, and $t(x)\in[0,1]$ is the medium transmission describing portion of scene radiance that reaches the camera.

Recovery of $J(x)$ requires estimation of $A$ and $t(x)$ from $I(x)$.

\begin{figure}[t]
  \centering
  \includegraphics[width=1\linewidth]{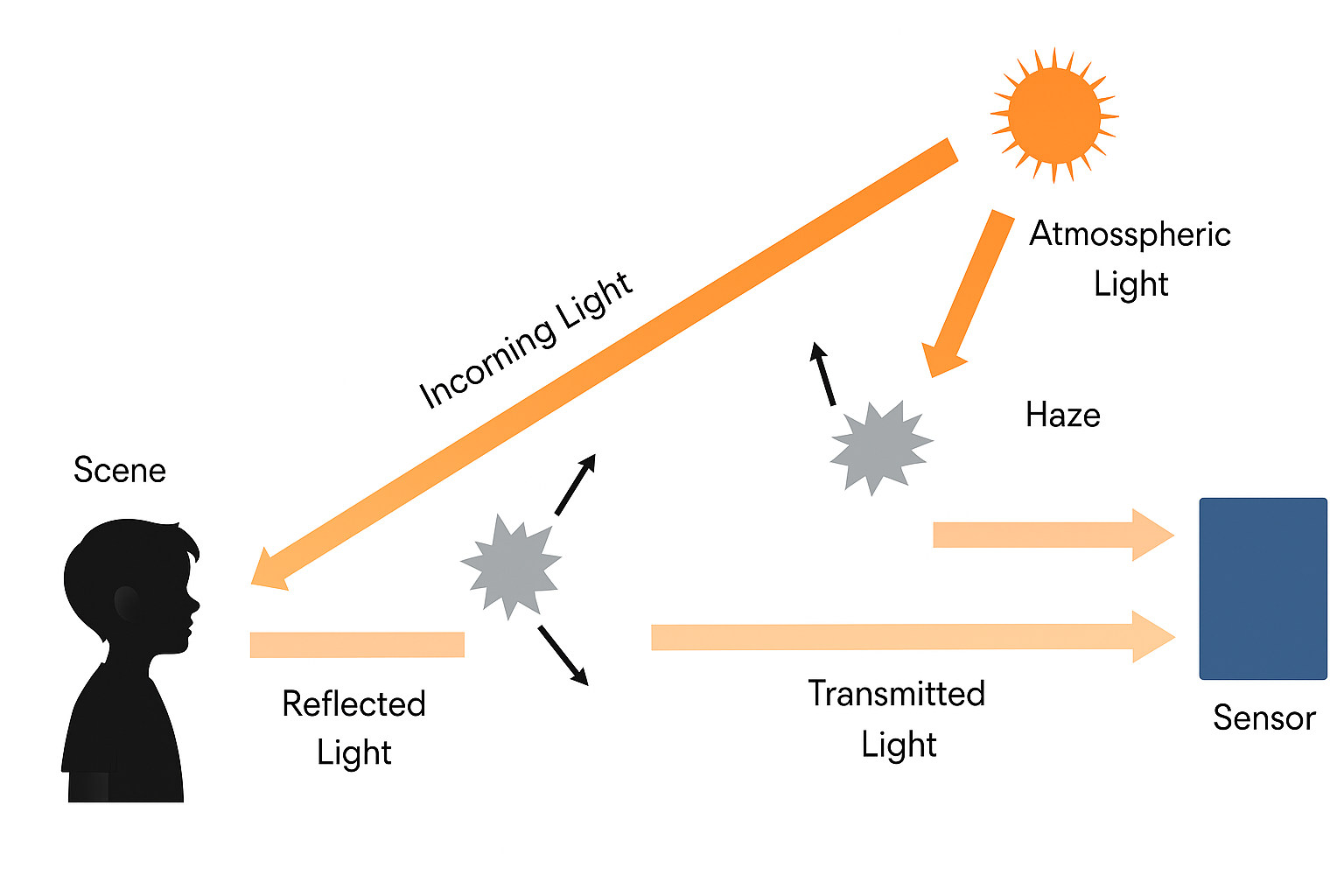}
  \includegraphics[width=1\linewidth]{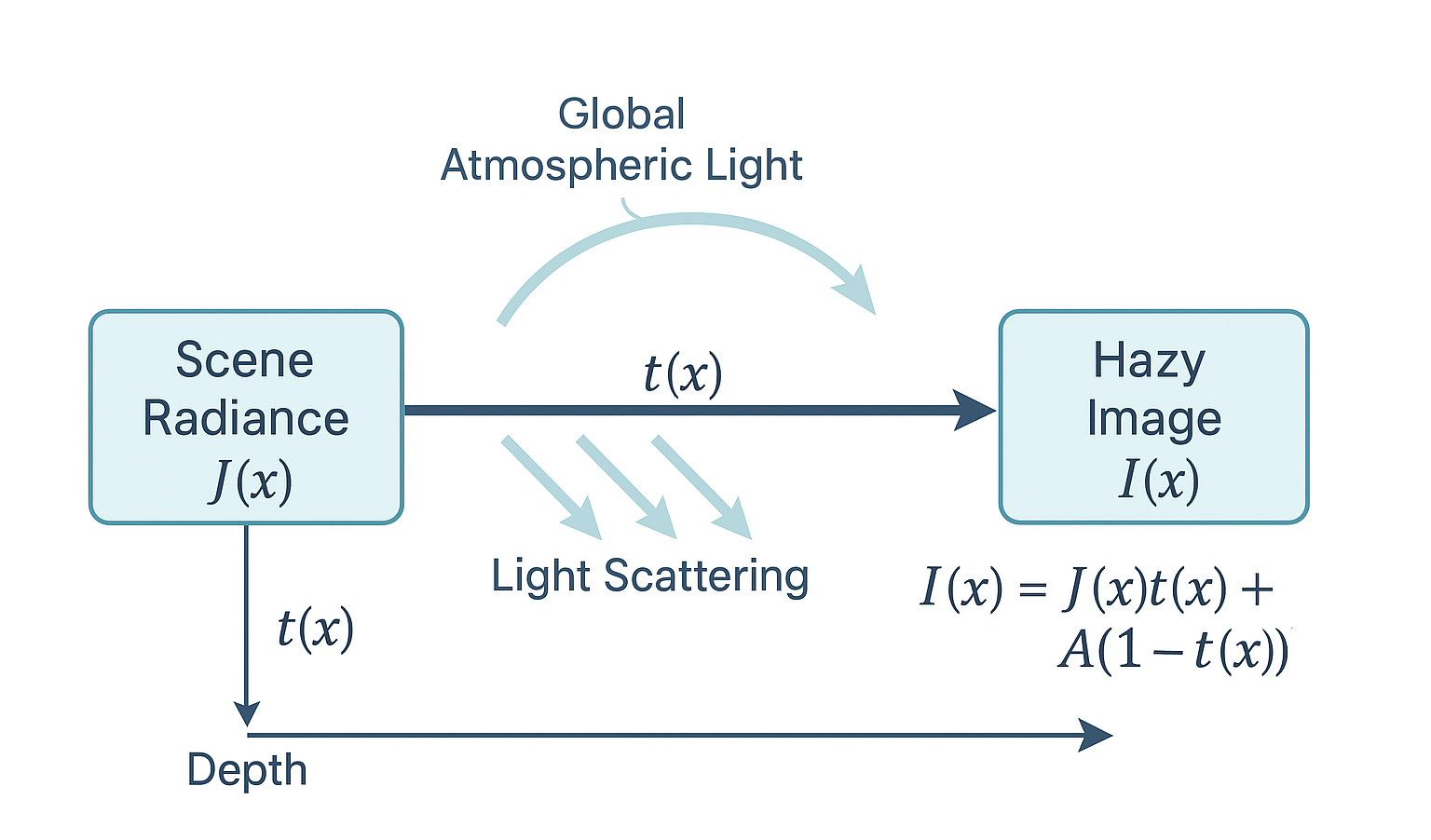}
  \caption{Atmospheric scattering model. $I(x)$ is the observed hazy image, $J(x)$ is the scene radiance, $A$ is atmospheric light and $t(x)$ is the transmission.}
  \label{fig:atmospheric-model}
\end{figure}

\section{Methodology}
Hazedefy uses a simplified version of the Dark Channel Prior (DCP) to achieve fast processing, along with an optional guided filter refinement for higher quality results on static images.

\subsection{Dark Channel Computation}
The dark channel $D(x)$ is computed as:
\begin{equation}
D(x) = \min_{y\in\Omega(x)} \big( \min_{c\in\{r,g,b\}} I^c(y) \big),
\end{equation}
where $\Omega(x)$ is a local patch (we use a $15\times15$ structuring element in the reference code). The implementation uses efficient morphological erosion on the per channel minimum to compute $D(x)$. 

\subsection{Atmospheric Light Estimation}

In the standard DCP approach, atmospheric light $A$ is estimated by selecting pixels with the highest values in the dark channel and using their RGB values. For video and real-time scenarios, this estimate can vary noticeably between frames. Hazedefy improves stability by selecting a fraction of the brightest dark channel pixels and averaging their RGB values, followed by scaling to prevent overestimation:

\begin{equation}
A = \alpha\cdot\frac{1}{K}\sum_{i=1}^{K} I(p_i), \qquad \alpha=0.8,
\end{equation}
where $p_i$ are the indices of the top $K$ dark channel pixels and $K$ is chosen as a fraction of total pixels (controlled by parameter $\omega$). This strategy, implemented in the reference dehaze module, reduces per frame jitter and produces more stable outputs across video frames. 

\subsection{Transmission Approximation}
A fast transmission approximation is computed as:
\begin{equation}
\hat t(x) = 1 - \omega\frac{\min_c I^c(x)}{\max(A)},
\end{equation}
then clamped with a lower bound $t_{\min}$ for numerical stability:
\begin{equation}
t(x) = \max\big(\hat t(x), t_{\min}\big).
\end{equation}
For the real time video mode empirical values $\omega=0.5$ and $t_{\min}=0.05$ are used in the shipped pipeline. These values were chosen to prioritize speed and avoid excessive amplification of noise in low transmission regions. 

\subsection{Guided Filter Refinement (Image Mode)}
For offline image mode, Hazedefy optionally applies a guided filter to the coarse transmission map to preserve edges while suppressing halos:
\begin{equation}
\tilde t = \text{GuidedFilter}(t, I_{\text{gray}}, r, \varepsilon),
\end{equation}
implemented following the efficient box-filter-based guided filter formulation \cite{he2013guided}. The image-mode parameters (radius $r$ and regularization $\varepsilon$) are set to produce artifact free reconstructions.

\subsection{Radiance Reconstruction and Post-processing}
The recovered scene radiance for channel $c$ is:
\begin{equation}
J^c(x) = \frac{I^c(x) - A^c}{t(x)} + A^c.
\end{equation}
After reconstruction the pipeline applies gamma correction:
\begin{equation}
J_\gamma(x) = 255\left(\frac{J(x)}{255}\right)^{1/\gamma},
\end{equation}
and an optional white-balance stage using OpenCV's \texttt{xphoto} module when available. These steps improve visual contrast and make colors appear more natural in many cases.

\section{System Design and Real-Time Considerations}
Hazedefy is built as a lightweight, frame-by-frame enhancement pipeline that focuses on consistent latency and efficient CPU usage. It follows a strictly feed forward design, which allows it to be used easily with live camera streams as well as offline video processing, without depending on information from previous frames.

Each input frame is first resized to a configurable resolution (by default $640\times480$) to keep the computational cost per frame under control. Dehazing is then performed using a simplified atmospheric scattering model derived from the Dark Channel Prior (DCP). All computations are implemented using efficient, vectorized operations, relying only on minimum filtering, simple normalization, and per pixel arithmetic. This avoids costly iterative optimization steps or learning based inference.

To improve temporal stability in video, atmospheric light is estimated by averaging a fraction of the brightest pixels in the dark channel, rather than selecting a single pixel. This approach reduces visible flickering between frames while remaining inexpensive to compute. For live video processing, guided filtering is deliberately excluded.

Overall, this design allows Hazedefy to achieve near real-time performance on consumer grade CPUs by combining reduced resolution processing, stable transmission estimation, and a minimal set of post processing steps, including gamma correction and optional color balancing.

\vspace{1mm}
\noindent\textbf{Algorithmic (per frame):}
\begin{enumerate}
  \item Acquire input frame and resize to target resolution.
  \item Compute per pixel channel minimum.
  \item Apply minimum filtering to obtain the dark channel.
  \item Estimate atmospheric light via fractional top dark channel averaging.
  \item Estimate transmission with a lower bound constraint.
  \item Recover scene radiance using the atmospheric scattering model.
  \item Apply gamma correction and optional color balance.
  \item Output the enhanced frame.
\end{enumerate}

The pipeline was evaluated on real world hazy images, recorded videos, and live camera feeds to verify perceptual improvement and real-time behavior under practical conditions.

\begin{figure}[t]
\centering
\begin{tikzpicture}[
  node distance=10mm,
  every node/.style={font=\footnotesize},
  arr/.style={-{Latex[length=3mm]}, line width=0.6pt},
  terminator/.style={draw, rounded corners=6pt, minimum width=40mm, minimum height=7mm, align=center},
  process/.style={draw, rectangle, rounded corners=2pt, minimum width=40mm, minimum height=7mm, align=center},
  io/.style={draw, trapezium, trapezium left angle=70, trapezium right angle=110, minimum width=40mm, minimum height=7mm, align=center},
  decision/.style={draw, diamond, aspect=2, inner ysep=0pt, minimum width=24mm, align=center}
]

\node[terminator] (start) {Start / Camera Input};
\node[io, below=of start] (capture) {Frame Capture / Resize (640×480)};
\node[process, below=of capture] (dark) {Dark Channel \\ (15×15 min filter)};
\node[process, below=of dark] (atmo) {Atmospheric Light Estimation \\ (top-$\omega$ avg)};
\node[process, below=of atmo] (trans) {Transmission Estimation \\ $t(x)=1-\omega\frac{\min(I)}{\max(A)}$};
\node[decision, below=of trans] (isimage) {Image Mode?};

\node[process, below right=8mm and 8mm of isimage] (guided)
    {Guided Filter \\ (edge-preserve)};

\node[process, below=24mm of isimage] (recover)
    {Radiance Recovery \\ $J^c=\frac{I^c-A^c}{t}+A^c$};

\node[process, below=of recover] (post)
    {Gamma Correction \\ \& Color Balance (opt.)};

\node[terminator, below=of post] (end) {Display / Save};

\draw[arr] (start) -- (capture);
\draw[arr] (capture) -- (dark);
\draw[arr] (dark) -- (atmo);
\draw[arr] (atmo) -- (trans);
\draw[arr] (trans) -- (isimage);

\draw[arr] (isimage.east) -- ++(10mm,0)
    node[midway, above, font=\scriptsize]{yes}
    -| (guided.north);

\draw[arr] (guided.south) |- (recover.east);

\draw[arr] (isimage.south) -- node[right=1mm, font=\scriptsize]{no} (recover.north);

\draw[arr] (recover) -- (post);
\draw[arr] (post) -- (end);

\end{tikzpicture}
\caption{Flowchart of Hazedefy system.}
\label{fig:hazedefy-flowchart}
\end{figure}
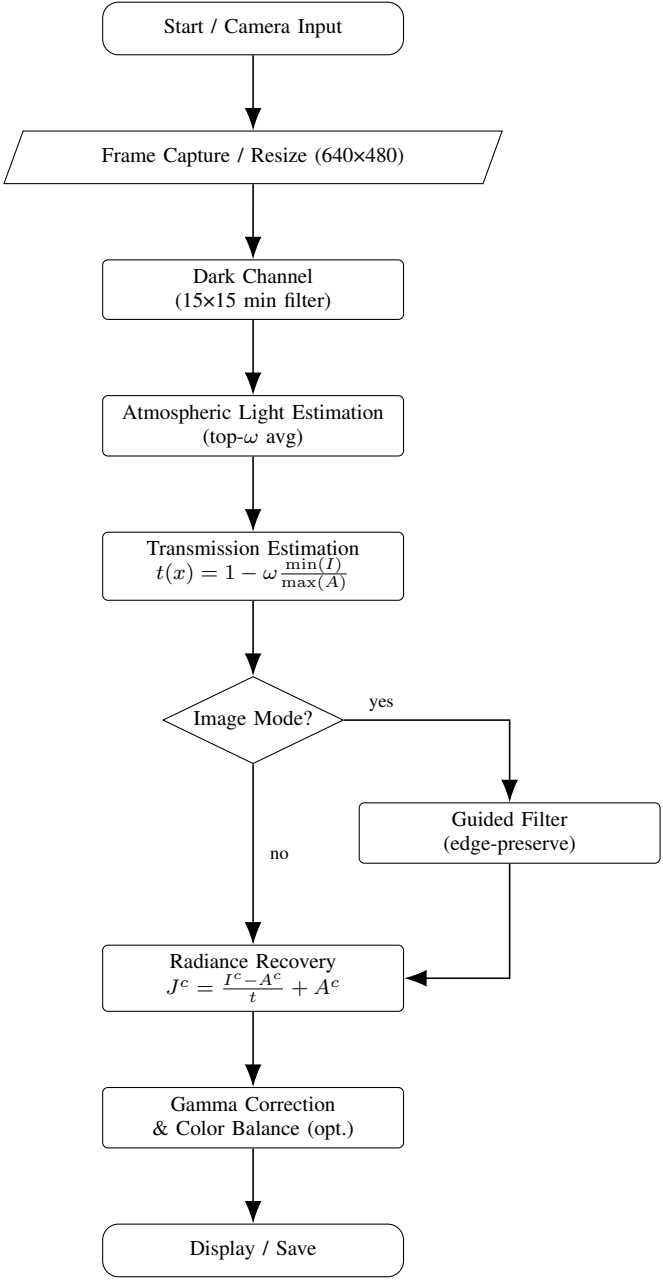





\section{Applications and Use Cases}
Hazedefy can be used in a wide range of vision systems operating in haze, fog, smoke, or other scattering media, particularly where real-time performance and low computational cost are required.

\noindent\textbf{A. Fire Safety and Emergency Response:}\\
\hspace*{1em}Real-time dehazing improves visibility in smoke filled environments, supporting navigation and awareness for firefighters, drones, and rescue robots.

\noindent\textbf{B. Satellite and Aerial Imaging:}\\
\hspace*{1em}Atmospheric scattering in satellite and UAV imagery can be reduced to improve clarity for remote sensing and environmental monitoring.

\noindent\textbf{C. Marine and Underwater Observation:}\\
\hspace*{1em}Dehazing enhances imagery in particle rich aquatic environments, helping underwater navigation and robotic exploration.

\noindent\textbf{D. Military and Defense Applications:}\\
\hspace*{1em}Better visibility helps systems perform surveillance, reconnaissance, and target identification more effectively in difficult or degraded conditions.

\noindent\textbf{E. Traffic Monitoring and Intelligent Transportation:}\\
\hspace*{1em}Dehazed video streams improve reliability of vehicle detection and traffic analysis in foggy and polluted scenes.

\noindent\textbf{F. Environmental Monitoring and Air-Quality Assessment:}\\
\hspace*{1em}Clearer imagery supports pollution monitoring and environmental surveillance.

\noindent\textbf{G. Aerial Drone Inspection:}\\
\hspace*{1em}Dehazing improves inspection imagery for infrastructure such as power lines, bridges, and wind turbines.



\section{Discussion and Limitations}
Hazedefy is intended for deployment scenarios where computational efficiency, algorithm transparency, and real-time operation are more important than the small visual gains offered by heavier learning based methods. Although the pipeline performs well in many practical situations, it also has some inherent limitations:
\begin{itemize}
\item Atmospheric light estimation can be affected in scenes with large bright regions, such as headlights or reflective surfaces, which may result in over enhanced outputs.
\item Performance may drop in scenes with complex or spatially varying lighting, or in cases of colored and non uniform haze, where a single scale prior is not sufficient.
\item The pipeline uses fixed, manually selected parameters and does not adapt automatically to different scenes. Exploring lightweight learning based refinements or scene adaptive parameter estimation is a possible direction for future work.
\end{itemize}

\section{Conclusion}
This paper presented Hazedefy, a lightweight image and video dehazing pipeline that combines a physics-based approach with practical engineering decisions to achieve near real-time performance on consumer-grade hardware. By focusing on simplicity, stability, and ease of deployment, the proposed method serves as an effective enhancement component for real-world vision systems operating under reduced visibility. The open and reproducible implementation also provides a useful reference for integration into mobile, embedded, and edge-based vision applications.

\section*{Acknowledgment} The Hazedefy implementation, documentation, and demonstration portal are publicly archived on Zenodo under DOI: 10.5281/zenodo.17915355. The web demo is hosted online.\footnote{https://hazedefy1.netlify.app/}.

\end{document}